\pdfoutput=1

\documentclass[11pt]{article}

\usepackage[final]{sty/acl}

\usepackage{times}
\usepackage{latexsym}

\usepackage[T1]{fontenc}

\usepackage[utf8]{inputenc}

\usepackage{microtype}

\usepackage{inconsolata}

\usepackage{graphicx}

\usepackage{booktabs}       
\usepackage{amsfonts}       
\usepackage{amssymb}
\usepackage{nicefrac}       
\usepackage{microtype}      
\usepackage{xcolor}         
\usepackage{amsmath}
\usepackage{amsfonts,amssymb}
\usepackage{multirow}
\usepackage{booktabs}
\usepackage{hyperref}
\usepackage{float}
\usepackage{enumitem}
\usepackage{algorithmicx}
\usepackage{algpseudocode}
\usepackage{tabularx}
\usepackage{subcaption}
\usepackage{marvosym}
\usepackage{textgreek}
\usepackage{wrapfig}
\usepackage{textcomp}
\usepackage{fontawesome}

\usepackage{amsmath,amsfonts,bm}

\def\eqref#1{equation~\ref{#1}}

\def\1{\bm{1}}

\def\vtheta{{\bm{\theta}}}

\def\vh{{\bm{h}}}

\def\vk{{\bm{k}}}

\def\vq{{\bm{q}}}

\def\vx{{\bm{x}}}

\DeclareMathAlphabet{\mathsfit}{\encodingdefault}{\sfdefault}{m}{sl}
\SetMathAlphabet{\mathsfit}{bold}{\encodingdefault}{\sfdefault}{bx}{n}

\def\sR{{\mathbb{R}}}

\def\emE{{E}}

\newcommand{\ci}{\mathrm{i}}

\newcommand{\benchmarkname}{\textsc{LongEmbed}}
\newcommand{\efbase}{E5\textsubscript{Base}}
\newcommand{\efrbase}{E5-RoPE\textsubscript{Base}}
\newcommand{\gtebase}{GTE\textsubscript{Base}}

\definecolor{darkblue}{rgb}{0, 0, 0.5}
\hypersetup{citecolor=darkblue}

\title{\textsc{LongEmbed}: Extending Embedding Models for Long Context Retrieval}

\newcommand{\snowflake}{\raisebox{-1pt}{\includegraphics[width=0.7em]{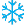}}}
\newcommand{\fire}{\raisebox{-1pt}{\includegraphics[width=0.7em]{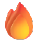}}}

\author{%
  Dawei Zhu~\thanks{~Contribution during Dawei's internship at MSR Asia. Sujian Li is the corresponding author.}$^{\heartsuit\spadesuit}$ \quad 
  Liang Wang~$^\diamondsuit$ \quad
  Nan Yang~$^\diamondsuit$ \quad 
  Yifan Song~$^{\heartsuit\spadesuit}$ \quad Wenhao Wu~$^{\heartsuit\spadesuit}$ \\ {\bf
  Furu Wei~$^\diamondsuit$  \quad Sujian Li~$^{\heartsuit\spadesuit\clubsuit}$} \\
  $^\heartsuit$~School of Computer Science, Peking University\\
  $^\spadesuit$~National Key Laboratory for Multimedia Information Processing, Peking University \\
  $^\clubsuit$~Jiangsu Collaborative Innovation Center for Language Ability, Jiangsu Normal University \\
  $^\diamondsuit$~Microsoft Corporation \\
  \texttt{\{dwzhu,lisujian\}@pku.edu.cn} \quad \texttt{wangliang@microsoft.com}\\
{\texttt{\url{https://github.com/dwzhu-pku/LongEmbed}}}
}

\begin{document}
\maketitle
\begin{abstract}

Embedding models play a pivotal role in modern NLP applications such as document retrieval.
However, existing embedding models are limited to encoding short documents of typically 512 tokens, restrained from application scenarios requiring long inputs.
This paper explores context window extension of existing embedding models, pushing their input length to a maximum of 32,768.
We begin by evaluating the performance of existing embedding models using our newly constructed  \benchmarkname{} benchmark, which includes two synthetic and four real-world tasks, featuring documents of varying lengths and dispersed target information.
The benchmarking results highlight huge opportunities for enhancement in current models.
Via comprehensive experiments, we demonstrate that training-free context window extension strategies can effectively increase the input length of these models by several folds.
Moreover, comparison of models using Absolute Position Encoding (APE) and Rotary Position Encoding (RoPE) reveals the superiority of RoPE-based embedding models in context window extension, offering empirical guidance for future models. Our benchmark, code and trained models will be released to advance the research in long context embedding models.

\end{abstract}

\section{Introduction}

Text embeddings are vector representations of natural language that encode its semantic information.
They play a pivotal role in various natural language processing (NLP) tasks, including information retrieval (IR) and retrieval-augmented generation (RAG).
However, embedding models for producing these vector representations still operates within a very narrow context window, many supporting only 512 input tokens~\citep{wang2022text,xiao2023c,ni2022large}. This narrow context window has greatly hindered their application in scenarios requiring long inputs, such as long Wikipedia articles and meeting scripts~\citep{saad2024benchmarking}.

Previous efforts that train a long context embedding model \textit{from scratch} suffer significant computational overhead, due to the combined demand for large batch sizes and long sequences.
For example, \citet{chen2024bge} utilized 96 A100 GPUs to train BGE-M3 which supports 8k context. Meanwhile, there have been many successes in extending context window of \textit{existing} LLMs in a plug-and-play way or via efficient fine-tuning, pushing their context from 4k to 128k~\citep{xiong2023effective} and even 2 million tokens~\citep{ding2024longrope}. Motivated by this, instead of training long context embedding models from scratch, this paper explores context window extension of \textit{existing} embedding models.

First, we examine the capability of existing embedding models in processing long context. Retrieval is selected as the proxy task, as it closely mirrors real-world application scenarios. While there have been some retrieval benchmarks such as BEIR~\citep{thakur2021beir} and LoCo~\citep{saad2024benchmarking}, we identify two major limitations with these existing benchmarks: 1) limited document length, 2) biased distribution of target information. To overcome this, we introduce the \benchmarkname{} benchmark that integrates two synthetic tasks to enable flexible control over document length, and four real tasks featuring dispersed target information. Results on \benchmarkname{} indicates huge room for improvement in current embedding models.

\begin{figure*}[t]
     \centering
      \begin{subfigure}[b]{0.30\textwidth}
         \centering
         \includegraphics[width=\textwidth]{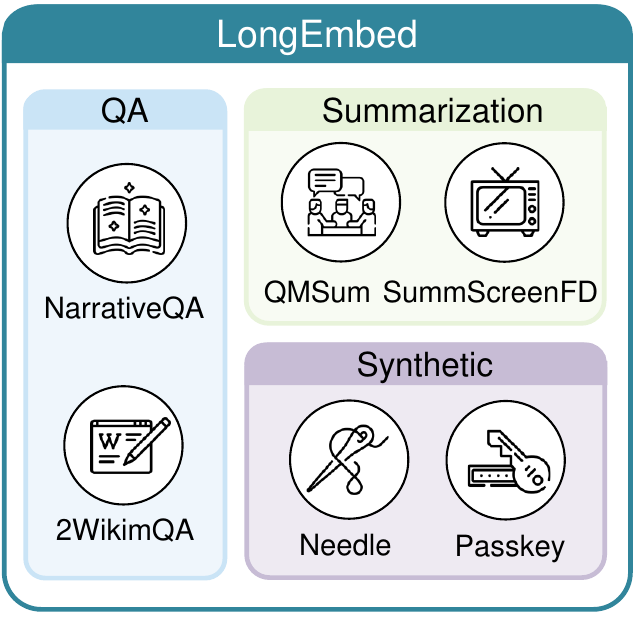}
         \caption{}
         \label{fig:long_embed}
     \end{subfigure}
     \hfill
     \begin{subfigure}[b]{0.36\textwidth}
         \centering
         \includegraphics[width=\textwidth]{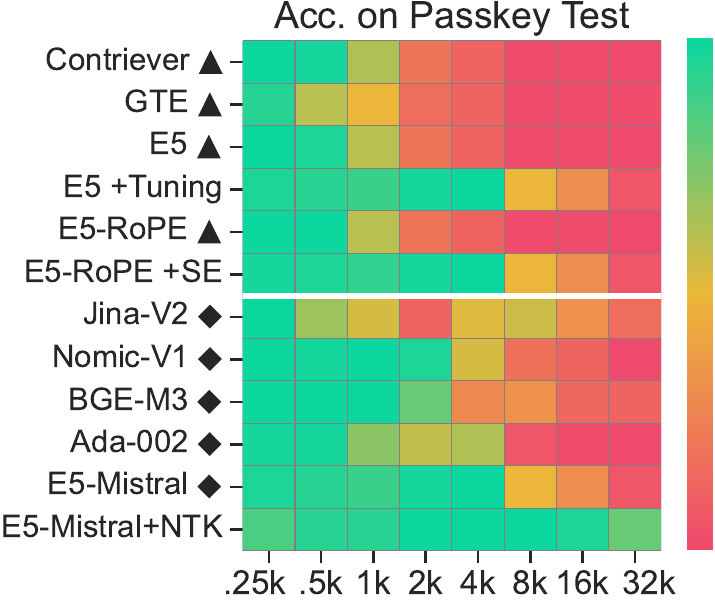}
         \caption{}
         \label{fig:benchmark_passkey_needle}
     \end{subfigure}
     \hfill
     \begin{subfigure}[b]{0.29\textwidth}
         \centering
         \includegraphics[width=\textwidth]{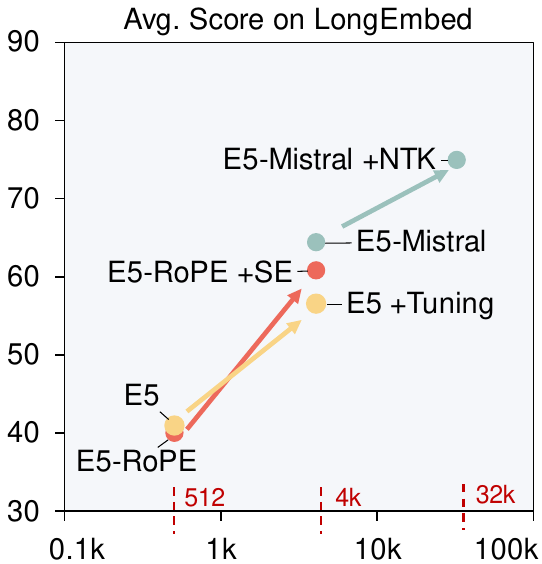}
         \caption{}
         \label{fig:overall_improvement}
     \end{subfigure}
     
     \caption{\textbf{(a)} Overview of the \benchmarkname{} benchmark. \textbf{(b)} Performance of current embedding models on passkey retrieval, with evaluation length ranging from 256 to 32,768~\protect\footnotemark{}.
     $\blacktriangle$ / $\blacklozenge$  denotes embedding models with 512 / $\ge$ 4k context. The greener \textcolor[HTML]{0cd79f}{\rule{0.7em}{0.7em}} a cell is, the higher retrieval accuracy this model achieves on the corresponding evaluation length. \textbf{(c)} Effects of context window extension methods on E5, E5-RoPE, E5-Mistral, measured by improvements of Avg. Scores on \benchmarkname{}. SE / NTK is short for SelfExtend / NTK-Aware Interpolation.}
     \label{fig:intro_fig}
\end{figure*}

Based on this, we explore plug-and-play strategies to extend embedding models, including parallel context windows, reorganizing position ids, and position interpolation. Comprehensive experiments show that these strategies can effectively extend the context window of existing embedding models by several folds, regardless of their original context being 512 or beyond 4k. Furthermore, for models employing absolute position encoding (APE), we show the possibility of harvesting further improvements via fine-tuning while strictly preserving original behavior within the short context. In this way, we have extended \efbase{}~\citep{wang2022text} from 512 to 4k~(See Figure~\ref{fig:overall_improvement}).

For models utilizing RoPE~\citep{su2021roformer}, substantial enhancements on \benchmarkname{} are observed when employing methods that fully leverage RoPE's advantages, such as NTK~\citep{ntk} and SelfExtend~\citep{jin2024llm}. 
As illustrated in Figure~\ref{fig:benchmark_passkey_needle} and~\ref{fig:overall_improvement}, leveraging NTK extends the context window of E5-Mistral to 32k, achieving close-to-perfect accuracy on passkey retrieval and state-of-the-art performance on \benchmarkname{}. Further, for fair comparison of APE / RoPE-based embedding models, we pre-train E5-RoPE following the training procedure and data of E5. Thorough comparison of E5 and E5-RoPE reveals the superiority of RoPE-based embedding models in context window extension.
To sum up, our contributions are as follows:
\begin{itemize}[leftmargin=*]
    \item We construct \benchmarkname{} to benchmark long context retrieval, which includes two synthetic and four real-world tasks, featuring documents of varying lengths and dispersed target information.
    \item We have conducted comprehensive experiments on training-free context window extension, extending the input length of existing embedding models by several folds. 
    \item We reveal the superiority of RoPE-based embedding models in context window extension via thorough comparison of models adopting APE and RoPE, offering empirical guidance for future embedding models.
    \item Our benchmark and trained models~(\efbase{}-4k, \efrbase{}) will be released to advance the research in long context embedding models.
\end{itemize}

\footnotetext{For simplicity, we report results from the \textit{base} versions of the included models by default. The supported context length of each model is presented in Table~\ref{tab:benchmark}. Inputs exceeding the supported context length are truncated.}

\section{Related Work}
\label{sec:related_work}
\noindent \textbf{Text Embedding Models.}\quad
Text embeddings encode semantic information of text as low-dimensional vectors, enabling numerous NLP applications. Early attempts on embeddings models include latent semantic indexing~\citep{deerwester1990indexing} and weighted average of word embeddings~\citep{mikolov2013efficient}. Modern embedding models~\citep{wang2022text,xiao2023c,neelakantan2022text} exploit supervision from labeled query-document pairs, adopting a multi-stage training paradigm that pre-trained on large-scale raw text pairs using contrastive loss, then fine-tuned on small scale but high-quality datasets.

Existing efforts in developing long-context embedding models typically involve first obtaining a long-context backbone model, either by pre-training with long inputs from scratch~\citep{gunther2023jina,nussbaum2024nomic,chen2024bge} or using existing ones~\citep{wang2023improving}, followed by training the backbone model to produce embeddings. Instead, this paper endows \textit{existing} embedding models with the ability to handle long context through context window extension.

\noindent \textbf{Context Window Extension for LLMs.}\quad
Due to the high cost of pre-training an LLM from scratch, there have been many efforts towards extending the context window of \textit{existing} LLMs in a plug-and-play manner. We categorize these efforts as follows: 1) \textit{Divide-and-conquer}, which involves segmenting long inputs into short chunks, processing each chunk with the model, and aggregating the results, as demonstrated by PCW~\citep{ratner2023parallel}; 2) \textit{Position reorganization}, which reorganizes position ids to accommodate longer inputs, as exemplified by SelfExtend~\citep{jin2024llm}, DCA~\citep{an2024training}. 3) \textit{Position interpolation}, which introduces new position embeddings by interpolating existing ones, includes PI~\citep{chen2023extending}, NTK~\citep{ntk}, YaRN~\citep{peng2023yarn}, and Resonance RoPE~\citep{wang2024resonance}. Our paper thoroughly investigates these three lines of methods on embedding models. We also acknowledge other efforts in context extension, such as token compression~\citep{jiang2023llmlingua,ge2023context,zhang2024soaring} and memory-based transformers~\citep{wang2024augmenting, xiao2024infllm}. However, the former is not applicable for bidirectional attention, and the latter requires complex mechanisms for accessing encoded content, hence we do not experiment with these two categories.

In addition to their plug-and-play usability, further fine-tuning on top of these methods with long training samples has been proven to yield better performance ~\citep{xiong2023effective,fu2024data,zhang2024extending,yen2024longcontext}. Addressing the overhead of training on long inputs and the scarcity of extremely long training data, a line of research investigates simulating long inputs within short context, including Randomized Positions~\citep{ruoss2023randomized}, Positional Skip-wise~(PoSE) training~\citep{zhu2023pose}, and SkipAlign~\cite{wu2024long}. 
This paper also leverage these efforts to synthesize long training samples from the original training data, facilitating further fine-tuning on top of plug-and-play methods.

\section{The \benchmarkname{} benchmark}

\begin{table*}[t]
\centering
\footnotesize
\renewcommand{\arraystretch}{1.1}
\setlength\tabcolsep{15pt}

\begin{tabular}{lccccc}
\toprule
\multirow{2.5}{*}{\textbf{Dataset}} & \multirow{2.5}{*}{\textbf{Domain}} & \multirow{2.5}{*}{\textbf{\# Queries}} & \multirow{2.5}{*}{\textbf{\# Docs}} & \textbf{Avg. Query} & \textbf{Avg. Doc} \\
& & & & \textbf{Words} & \textbf{Words} \\\midrule
\multicolumn{6}{c}{\textit{Real Tasks}} \\ \midrule
NarrativeQA & Literature, Film & 10,449 & 355 & 9 & 50,474  \\
QMSum & Meeting & 1,527 & 197 & 71 & 10,058  \\
2WikiMultihopQA & Wikipedia & 300 & 300 & 12 & 6,132 \\
SummScreenFD & ScreenWriting & 336 & 336 & 102 & 5,582 \\ \midrule
\multicolumn{6}{c}{\textit{Synthetic Tasks}} \\ \midrule
Passkey & Synthetic & 400 & 800 & 11 & \dag{} \\
Needle & Synthetic & 400 & 800 & 7 & \dag{} \\
\bottomrule
\end{tabular}
\caption{Overview of the \benchmarkname{} benchmark. Average word number is rounded to the nearest integer. \dag{} For needle and passkey test, we have 8 groups of queries and candidate documents, with the documents averaging $\{0.25,0.5,1,2,4,8,16,32\}\times 0.75k$ words, respectively.}
\label{tab:statistics}
\end{table*}
In this section, we first identify two limitations of existing retrieval benchmarks for evaluating long-context capabilities~(\S~\ref{sec:cons_ext_bench}). Then, we introduce the retrieval tasks adopted in our \benchmarkname{}, including both synthetic ones~(\S~\ref{sec:syn_tasks}) and real ones~(\S~\ref{sec:real_tasks}). 

\subsection{Examing Existing Retrieval Benchmarks}
\label{sec:cons_ext_bench}

There are two main desiderata for curating a long context retrieval benchmark. First, the candidate documents should be long enough. Second, the target information to answer user query should be as uniformly distributed across the document as possible. This prevents embedding models from solely focusing on specific parts, such as the beginning~\citep{coelho2024dwell}, to achieve unreasonably high scores.
Based on these criteria, we examine existing retrieval benchmarks as follows:
\begin{figure}[t]
    \centering
    \includegraphics[width=0.45\textwidth]{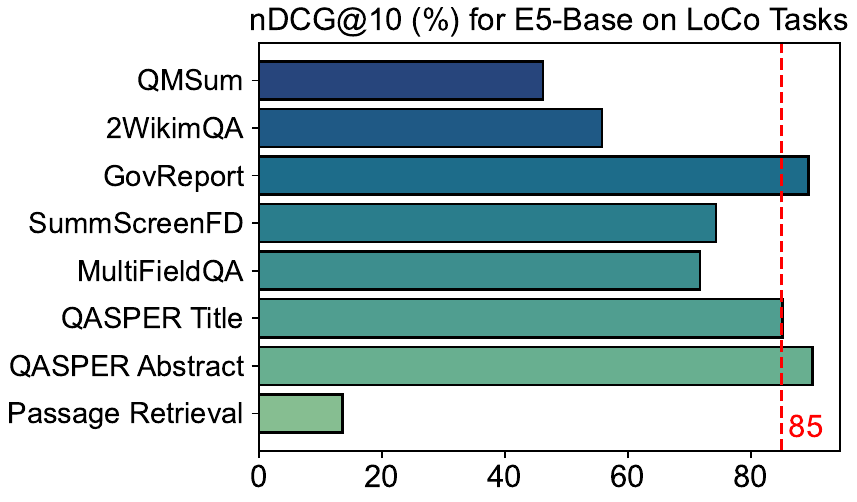}
    \caption{Results of \efbase{} on 8 LoCo tasks that are publicly available. }
    \label{fig:loco_ndcg}
\end{figure}

\noindent \textbf{BEIR Benchmark}~\citep{thakur2021beir} is a collection of 18 information retrieval datasets, ranging across ad-hoc web search, question answering, fact verification, etc. However, documents in this benchmark contains fewer than 300 words on average~(See Table~\ref{tab:beir} in Appendix), making it unsuitable for measuring long context retrieval that usually involves documents of thousands or tens of thousands of words. 


\noindent \textbf{LoCo Benchmark}~\citep{saad2024benchmarking} consists 12 retrieval tasks that requires long context reasoning, spanning diverse domains such as law and finance. However, it still suffers from biased distribution of key information, as demonstrated in Figure~\ref{fig:loco_ndcg}. With only 512 context length, \efbase{} achieves >85\% nDCG scores on 3 out of 8 publicly-available LoCo tasks. This severely biased distribution of target information undermines its ability to reflect model performance as context increases.

\subsection{Synthetic Tasks in \benchmarkname{}}
\label{sec:syn_tasks}

First, we introduce the passkey and needle retrieval task for embedding models as follows:

\noindent \textbf{Personalized Passkey Retrieval.}\quad Passkey retrieval~\citep{mohtashami2023landmark} requires LLMs to recover a random passkey hidden within a long document comprising garbage information. 
For embedding models, we adopt the \textit{personalized passkey retrieval}~\citep{wang2023improving}, where each document contains a unique person name and his/her passkey at random position. The goal is to retrieve the document containing the given person’s passkey from all candidates documents.

\begin{figure}[t]
    \centering
    \includegraphics[width=0.49\textwidth]{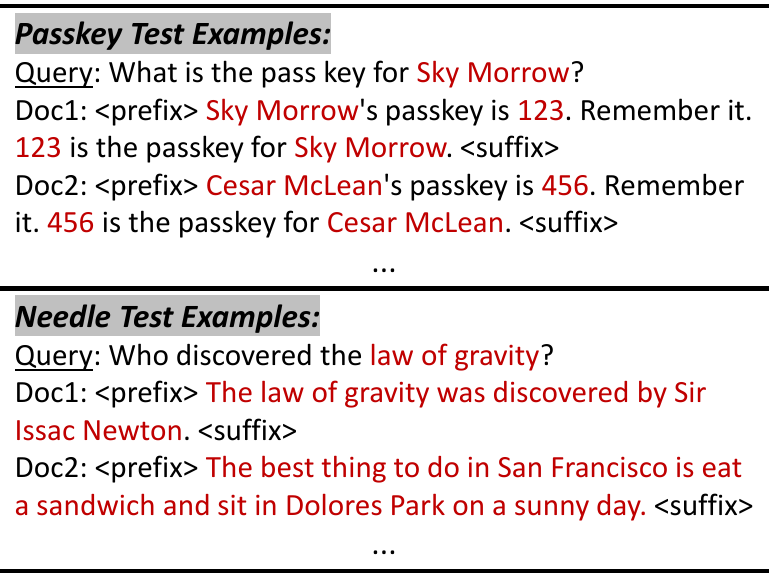}
    \caption{Example for the passkey and needle test. For the passkey test, the \textit{<prefix / suffix>} are repeats of \textit{"The grass is green. The sky is blue. The sun is yellow. Here we go. There and back again."} For the needle test, the \textit{<prefix>} and \textit{<suffix>} form a long essay.}
    \label{fig:passkey_needle_example}
\end{figure}

\noindent \textbf{Needle-in-a-haystack Retrieval.}\quad
While passkey retrieval surrounds key information with garbage sentences, needle-in-a-haystack retrieval~\citep{needleinhaystack,liu2024lost} randomly inserts key information into an arbitrary position of a long essay, making the task more challenging.
To tailor this task for embedding models, we instruct GPT-4 to generate 100 facts covering a variety of domains including physics, history, geometry, art, etc, and 100 
\textit{queries} correspondingly. The facts are subsequently treated as \textit{needles} and randomly inserted into the PaulGrahamEssay to form 100 candidate documents.  Our task is to correctly retrieve the document that contains corresponding needle given the query. 

The advantage of synthetic data is that we can flexibly control context length and distribution of target information. For both tasks, we evaluate a broad context range of $\{0.25,0.5,1,2,4,8,16,32\}\times1,024$ tokens~\footnote{Since token numbers vary w.r.t. tokenizers, we use a rough estimation that 1 token = 0.75 word, and constraint the word numbers to not exceed $\{0.25,0.5,1,2,4,8,16,32\}\times1,024\times0.75$.}. For each context length, we include 50 test samples, each comprising 1 query and 100 candidate documents.~\footnote{The original version of personalized passkey retrieval uses different candidate documents for each query, resulting in 50 queries and 5,000 documents to encode for each context length. To speed up evaluation, we share the candidate documents for different queries within each context length.}  In this way, we can measure the effective context size of embedding models for up to 32k tokens. Examples for both tasks are in Figure~\ref{fig:passkey_needle_example}.

\subsection{Real Tasks in \benchmarkname{}}
\label{sec:real_tasks}
While synthetic tasks offer flexibility in manipulating context length and distributing target information, they still differ from real-world scenarios. 
To conduct a comprehensive evaluation, we have tailored following long-form QA and summarization tasks for long context retrieval. For QA datasets, we use the questions as queries, the set of all input documents as candidate documents. For summarization datasets, we use the summaries as queries, and the set of all input documents as candidate documents.

\noindent \textbf{NarrativeQA}~\citep{kocisky-etal-2018-narrativeqa} is a QA dataset comprising long stories and corresponding questions about specific content such as characters, events. As these details are dispersed throughout the story, models must process the entire long context to get the correct answers.

\noindent \textbf{2WikiMultihopQA}~\citep{xanh2020_2wikimultihop} is a multi-hop QA dataset featuring questions with up to 5 hops, synthesized through manually designed templates to prevent shortcut solutions. This necessitates the ability to process and reason over long context, ensuring that answers cannot be obtained by merely focusing on a short span within the document. 

\noindent \textbf{QMSum}~\citep{zhong2021qmsum} is a query-based meeting summarization dataset that requires selecting and summarizing relevant segments of meetings in response to queries. Due to the involvement of multiple participants and topics in the meeting, summarization regarding specific queries naturally requires aggregating information dispersed throughout the entire text. 

\noindent \textbf{SummScreenFD}~\citep{chen2022summscreen} is a screenplay summarization dataset comprising pairs of TV series transcripts and human-written summaries. Similar to QMSum, its plot details are scattered throughout the transcript and must be integrated to form succinct descriptions in the summary.

Table~\ref{tab:statistics} presents the overall statistics of \benchmarkname{}. Considering the computational complexity that increases quadratically with input length, we intentionally restrict the number of candidate documents in each task to to not exceed $10^3$. In this way, we can efficiently evaluate the basic long context capabilities of embedding models. For further elaboration on the source and examples for each dataset, please refer to Appendix~\ref{app:sec_longembed_details}.

\section{Methodology}

\subsection{Preliminary: APE \& RoPE}
\noindent \textbf{Absolute Position Embedding (APE)} stands as the predominant positional encoding strategy for embedding models, as majority of them follows the BERT architecture~\citep{devlin-etal-2019-bert}. APE-based models first embed absolute position ids into position vectors and add token embeddings to their corresponding position vectors, before feeding them to a stack of transformer layers. 

\noindent \textbf{Rotary Position Embedding (RoPE)} is the most pervasive position embedding strategy in the era of LLMs, including LLaMA~\citep{touvron2023llama},  QWen~\citep{bai2023qwen}, etc. It encodes position information of tokens with a rotation matrix that naturally incorporates explicit relative position dependency. To elucidate, given a hidden vector $\vh=[h_0,h_1,...,h_{d-1}]$ of dimension $d$, and a position index $m$, RoPE operates as follows: 
\begin{align*}
    f(\vh,m) &= [(h_0+\ci h_1)e^{\ci m\theta_0}, (h_2+\ci h_3)e^{\ci m\theta_1},...,\\&(h_{d-2}+\ci h_{d-1})e^{\ci m\theta_{d/2-1}}]
\end{align*}
where $\theta_j=10000^{-2j/d},j\in\{0,1,...,d/2-1\}$, $\ci=\sqrt{-1}$ is the imaginary unit. Unlike APE that is directly applied to the input vector $\vx$, RoPE is employed on the query and key vectors at each layer. The attention score $a(\vq,\vk)$ between a query $\vq$ at position $m$ and a key $\vk$ at position $n$ is:
\begin{equation}
\label{eqn:attention_score}
\resizebox{1.0\linewidth}{!}{$
\begin{aligned} 
a&(\vq,\vk)=\mathrm{Re}\langle f(\vq,m), f(\vk,n) \rangle \\
&=\mathrm{Re}\left[\sum_{j=0}^{d/2-1}(q_{2j}+\ci q_{2j+1})(k_2j-\ci k_{2j+1})e^{\ci (m-n)\theta_j}\right] \\
&:=g(\vq,\vk,(m-n)\vtheta) 
\end{aligned}
$}
\end{equation}
where g(·) is an abstract mapping function exclusively dependent on $\vq,\vk$ and $(m-n)\vtheta$.

\subsection{Extending APE-based Models}

As delineated in Section~\ref{sec:related_work}, training-free context extension strategies applicable to embedding models can be classified into 3 categories: 1) Divide-and-conquer; 2) Position reorganization; 3) Position interpolation. In this section, we introduce methods from each of these categories to assess their applicability to embedding models. Further fine-tuning on top of these methods is also included. Let $L_o$ represent the original context length, $\mathcal{D} = \{x_1,x_2,...,x_{L_t}\}$ denote a long document of target context length $L_t$, and $s = \lceil L_t/L_o \rceil$ indicate the context scaling factor. The context extension methods we investigated are described below:

\begin{figure*}[t]
    \centering
    \includegraphics[width=0.9\textwidth]{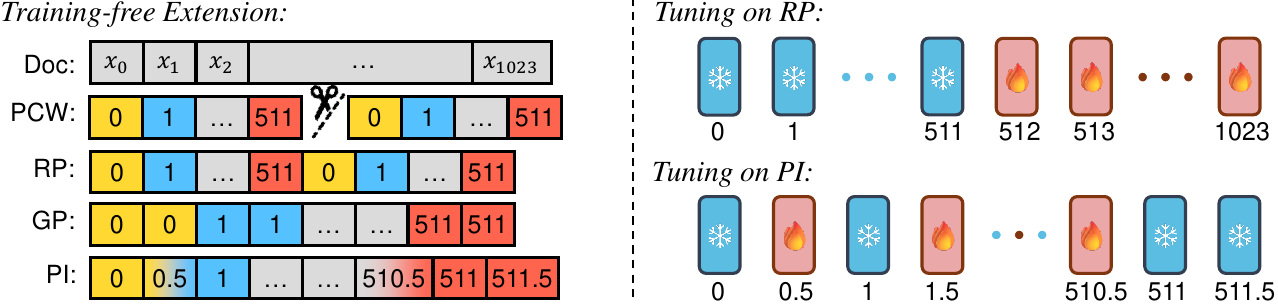}
    \caption{(Left) Arrangement of pids for extending APE-based models from 512 to 1,024. (Right) Illustration of learnable~(\fire{}) and frozen~(\snowflake{}) position vectors when further tuning on RP / PI.}
    \label{fig:ape_demo}
\end{figure*}

\noindent \textbf{Parallel Context Windows (PCW).}\quad To process a long document with a short-context model, PCW divides the long document into multiple short chunks, processes each chunk in parallel, and aggregates their results~\citep{ratner2023parallel,yen2024longcontext}. In practice, we first segment $\mathcal{D}$ into chunks of $L_o$ tokens, then average over each chunk's embeddings to represent $\mathcal{D}$. For simplicity, we set the overlap between adjacent chunks to 0, except for the last chunk, to ensure it contains $L_o$ tokens.

\noindent \textbf{Grouped \& Recurrent Positions (GP \& RP).}\quad Dividing inputs into chunks and processing them separately sacrifices their interaction in between. By contrast, position reorganization accommodates longer context by reusing the original position ids. To be specific, we experiment with two simple strategies: \textit{Grouped Positions} and \textit{Recurrent Positions}. The former groups the original position ids as such: $f_{gp}(pid) \rightarrow \lfloor pid / s \rfloor$, while the latter assigns the position ids recurrently,
formulated as:  $f_{rp}(pid) \rightarrow pid\bmod L_o$.

\noindent \textbf{Linear Position Interpolation (PI).}\quad Instead of reusing position ids, \citet{chen2023extending} introduces new position embeddings via linear interpolation of existing ones. To apply PI on APE-based models, we map the positions ids as such: $f_{pi}(pid) \rightarrow pid / s$, and assign embeddings for non-integers as linear interpolation of that of neighboring integers. In practice, we first extend the original position embedding matrix $\emE_o \in \sR^{L_o\times d}$ into $\emE_t \in \sR^{L_t\times d}$, where $d$ stands for hidden size. Next, we assign $\emE_t[i\cdot s]=\emE_o[i], i\in\{0,1,...,L_o-1\}$. For non-integer position id $j$ between $i$ and $i+1$, we determine their embeddings as follows: $\emE_t[s\cdot j] = ((i+1-j)\emE_t[i\cdot s] + (j-i)\emE_t[(i+1)\cdot s])$.

\noindent \textbf{Further Tuning.}\quad Except for PCW, which divides long texts into smaller blocks and processes separately, GP, RP, and PI can all be seen as extending the position embedding matrix. Since APE-based models assign an independent vector to each position, we can freeze the original model parameters while updating only the newly added position embeddings. In this way, we can strictly maintain model ability within 512 context, while harvesting further performance gains in handling long context as free lunch. Specifically, further fine-tuning on top of RP and PI is explored in this paper, as illustrated in Figure~\ref{fig:ape_demo}~(Right). Since the traditional training data for embedding models are short queries and passages not exceeding 512 tokens, we manipulate position ids to simulate long training samples, as proposed in~\citet{zhu2023pose}. See Appendix~\ref{sec:app_implementation} for details of further fine-tuning.

\subsection{Extending RoPE-based Models}

For RoPE-based models, we further explore Self Extend and NTK, which respectively advances over GP and PI, harnessing the inherent advantages of RoPE. Since there is no simple strategy for further training while exactly maintaining original performance like APE, we leave comprehensive exploration of training-based context window extension for RoPE-based models for future work.

\noindent \textbf{Self Extend~(SE).}\quad
Compared with APE, RoPE operates on the query and key vectors at each layer to encode relative positions, offering enhanced flexibility for position reorganization. For each token, instead of assigning grouped relative positions to all other tokens, SelfExtend~\citep{jin2024llm} re-introduces normal relative positions within the nearest neighbor window $w$, achieving improved performance. For example, consider a document of 10 tokens $\{x_0,x_1,...,x_9\}$ with a neighbor window size $w=4$ and a group size $g=2$. The relative positions to $x_0$ are $\{0, 1, 2, 3, 4, 4, 5, 5, 6, 6\}$. For $x_4$, the relative positions of the other tokens are $\{-4, -3, -2, -1, 0, 1, 2, 3, 4, 4\}$.

\begin{table*}[t]
\centering
\footnotesize
\renewcommand{\arraystretch}{1.1}
\setlength\tabcolsep{5pt}

\begin{tabular}{lrcccccccc}
\toprule
\multirow{2.5}{*}{\textbf{Model}} & \multirow{2.5}{*}{\textbf{Param.}} & \multirow{2.5}{*}{\textbf{CTX Len.}} &
\multicolumn{2}{c}{\textbf{Synthetic (Acc@1)}} & \multicolumn{4}{c}{\textbf{Real (nDCG@10)}} & \multirow{2.5}{*}{\textbf{Avg.}} \\ \cmidrule(r){4-5} \cmidrule(r){6-9}
& & & \textbf{Passkey} & \textbf{Needle} & \textbf{NQA} & \textbf{QMS} & \textbf{SFD} & \textbf{WQA} &  \\ \midrule
\multicolumn{10}{c}{\textit{512 Context Models}} \\ \midrule
\efbase{}~\citep{wang2022text} & 110M & 512 & 38.0 & 28.5 & 25.3 & 23.8 & 74.7 & 55.8 & \textbf{41.0} \\
\efrbase{} & 110M & 512 & 38.5 & 31.5 & 24.6 & 23.2 & 66.6 & 58.8 & 40.5 \\
\gtebase~\citep{li2023towards} & 110M & 512 & 31.0 & 24.5 & 28.6 & 21.8 & 55.8 & 47.3 & 34.8 \\ 
BGE\textsubscript{Base}~\citep{xiao2023c} & 110M & 512 & 18.0 & 25.3 & 25.6 & 22.4 & 60.3 & 51.7 & 33.9 \\ 
Contriever~\citep{izacard2021towards} & 110M & 512 & 38.5 & 29.0 & 26.7 & 25.5 & 73.5 & 47.3 & 40.1 \\ 
GTR\textsubscript{Base}~\citep{ni2022large} &  110M & 512 & 38.5 & 26.3 & 26.5 & 18.3 & 63.7 & 52.2 & 36.5 \\ 
\midrule
\multicolumn{10}{c}{\textit{$\ge$ 4k Context Models}} \\ \midrule
E5-Mistral~\citep{wang2023improving} & 7B & 4,096 & 71.0 & 48.3 & 44.6 & 43.6 & 96.8 & 82.0 & \textbf{64.4} \\ 
Jina-V2~\citep{gunther2023jina} & 137M & 8,192 & 50.3 & 54.5 & 37.9 & 38.9 & 93.5 & 74.0 & 58.2 \\ 
Nomic-V1\citep{nussbaum2024nomic} & 137M & 8,192 & 60.7 & 39.5 & 41.2 & 36.7 & 93.0 & 73.8 & 57.5 \\ 
BGE-M3~\citep{chen2024bge} & 568M & 8,192 & 59.3 & 40.5 & 45.8 & 35.5 & 94.0 & 78.0 & 58.9 \\
OpenAI-Ada-002 & - & - & 50.8 & 36.8 & 41.1 & 40.0 & 91.8 & 80.1 & 56.8 \\ 
\midrule
\multicolumn{10}{c}{\textit{Our Extended Models}} \\ 
\midrule
\efbase{} + Tuning (4k) & 110M & 4,096 & 67.3 & 41.5 & 30.4 & 35.7 & 95.2 & 69.2 & 56.6 \\
\efrbase{} + SelfExtend (4k) & 110M & 4,096 & 73.5 & 53.5 & 32.3 & 39.1 & 91.9 & 74.6 & 60.8  \\
E5-Mistral + NTK (32k) & 7B & 32,768 & \textbf{93.8} & \textbf{66.8} & \textbf{49.8} & \textbf{49.2} & \textbf{97.1} & \textbf{95.2} & \textbf{75.3} \\
\bottomrule
\end{tabular}
\caption{Results~(\%) of existing and extended embedding models on \benchmarkname{}. \textit{NQA}, \textit{QMS}, \textit{SFD}, \textit{WQA} is short for \textit{NarrativeQA}, \textit{QMSum}, \textit{SummScreenFD}, \textit{2WikiMultihopQA}, respectively. We show that context window extension can effectively improve existing embedding models in processing long context.}
\label{tab:benchmark}
\end{table*}

\noindent \textbf{NTK-Aware Interpolation~(NTK).}\quad Given a scaling factor $s$, PI proportionally down-scales position index $m$ to $m/s$. In this way, the attention score $a(\vq,\vk)$ defined in Equation~\ref{eqn:attention_score} becomes $g(\vq,\vk,(m-n)\vtheta/s)$. This is also equivalent to reducing the frequencies $\vtheta$ uniformly, which may prevent the model from learning high-frequency features, as shown by the  Neural Tangent Kernel (NTK) theory~\citep{jacot2018neural}. To remedy this, NTK-Aware interpolation~\citep{ntk} scales high frequencies less and low frequencies more to spread out the interpolation pressure across multiple dimensions. This is achieved by directly altering  the original $\theta_j=10000^{-2j/d}$ into $\theta'_j=(10000\lambda)^{-2j/d}$, where $\lambda$ is conventionally chosen to be slightly greater than $s$.

\section{Experiments}

\subsection{Experimental Setup}

\noindent \textbf{Benchmarked Models.}\quad We evaluate both open-sourced and proprietary models on \benchmarkname{}, including E5, GTE, BGE, Contriever, GTR, E5-Mistral, Jina-V2, Nomic-V1, BGE-M3, OpenAI-ada-002. 
M2~\citep{saad2024benchmarking} is not included in our evaluation, given its training data partly overlaps with test samples in \benchmarkname{}.

\noindent \textbf{Candidate Models for Extension.}\quad 
From each of the APE-based and RoPE-based category, we select 2 candidate models for comprehensive study. The former includes \efbase{} and \gtebase. The latter includes the 4,096-context E5-Mistral, and a newly trained \efrbase{}, which supports 512 context~(See Appendix~\ref{sec:app_train_e5rope} for its training details and BEIR results). Note that \efrbase{} employs the same training procedure and training data as \efbase{}, only with APE substituted with RoPE. This facilitates fair comparison of APE / RoPE-based models in context window extension, as presented in Section~\ref{sec:analysis}. For implementation details of each context window extension strategies on each model, please refer to Appendix~\ref{sec:app_implementation}.


\subsection{Main Results}

     


Table 2 demonstrates the performance of existing embedding models on our \benchmarkname{} benchmark. Among the 512-context models, \efbase{} achieves the highest average score of 41.0 points, closely followed by \efrbase{} and Contriever. As the supported context length increases beyond 4k, exemplified by E5-Mistral and Jina-V2, a discernible increase in scores is observed. This verifies both the efficacy of these long-context models and the validity of \benchmarkname{} to assess long-context retrieval. Note that even the best performing model attains only 64.4 pts on average, indicating huge room for improvement in current models.

In the last row block of Table 2, we further include the best results achieved by \efbase{}, \efrbase{} and E5-Mistral after context window extension. For \efbase{} and \efrbase{}, we extend their contexts from 512 to 4,096. For E5-Mistral, we extend its context from 4,096 to 32,768. Compared to the original versions, the extended models achieve an average score increase of +15.6 / +20.3 / +10.9 points. This indicates the efficacy of these context extension strategies on embedding models, enabling them to handle inputs of several folds longer. Detailed performance comparison of different extension strategies on APE \& RoPE-based embedding models is presented in Section~\ref{sec:ext_comp}.

\subsection{Comparison of Extension Methods}
\label{sec:ext_comp}

\begin{figure}
    \centering
    \includegraphics[width=\linewidth]{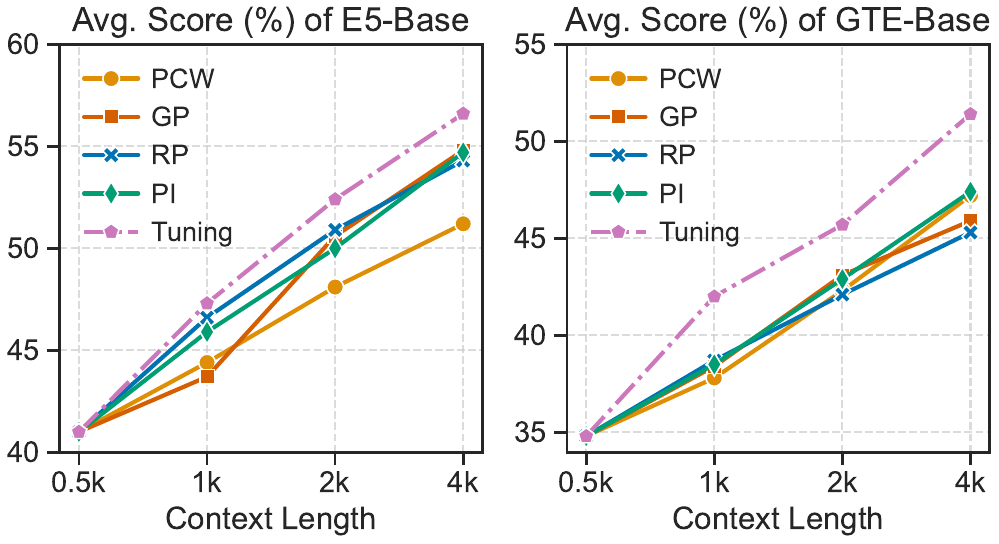}
    \caption{Effects of different context window extension methods on \efbase{} and \gtebase. We show that further tuning yields the best results.}
    \label{fig:e5_gte_ape}
\end{figure}

\noindent \textbf{APE-based Models.}\quad 
Figure~\ref{fig:e5_gte_ape} illustrates the impact of various context extension strategies on \efbase{} and \gtebase{} across different target context lengths. We observe that plug-and-play methods including GP, RP, PI and PCW strategies yield comparable results with no significant disparities. On the other hand, further tuning consistently yields additional performance gains for both models, across all target context lengths. Particularly noteworthy is \gtebase, which showcases a substantial average score increase of approximately 5 points after further tuning. This suggests that freezing the original model weights and fine-tuning exclusively the added position embeddings can effectively extend the model's context window while strictly maintaining model's original ability.


\begin{figure}[t]
     \centering
     \begin{subfigure}[b]{0.49\linewidth}
         \centering
         \includegraphics[width=\linewidth]{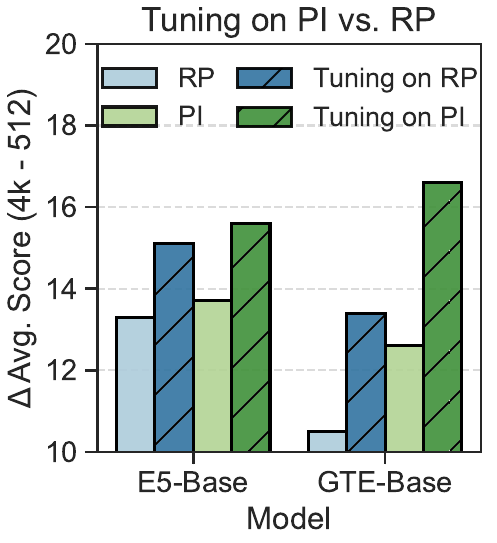}
         \caption{}
         \label{fig:pi_vs_rp}
     \end{subfigure}
     \hfill
     \begin{subfigure}[b]{0.49\linewidth}
         \centering
         \includegraphics[width=\linewidth]{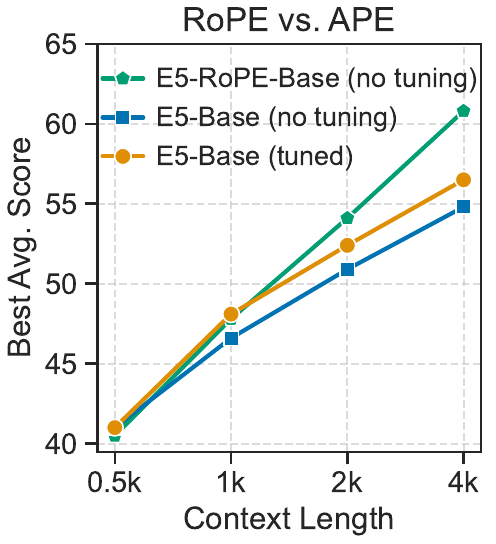}
         \caption{}
         \label{fig:rope_vs_ape}
     \end{subfigure}
     
     \caption{(a) Performance gain after tuning on PI / RP, compared with the original model. (b) Best results achieved by
extended versions of \efbase{} / \efrbase{}.}
     \label{fig:passkey_needle}
\end{figure}

\begin{table}[t]
\centering
\footnotesize
\renewcommand{\arraystretch}{1.1}
\setlength\tabcolsep{2.5pt}
\begin{tabular}{lccccccc}
\toprule
\multirow{2.5}{*}{\textbf{Model}} &
\multicolumn{2}{c}{\textbf{Synthetic}} &
\multicolumn{4}{c}{\textbf{Real}} & \multirow{2.5}{*}{\textbf{Avg.}}  \\
\cmidrule(r){2-3} \cmidrule(r){4-7}
& P & N & NQA & QMS & SFD & WQA & \\
\midrule
\textit{\efrbase{}} & \textit{38.5} & \textit{31.5} & \textit{24.6} & \textit{23.2} & \textit{66.6} & \textit{58.8} & \textit{40.5} \\ \midrule
+PCW (4k) & 42.5 & 50.8 & 25.1 & 34.9 & \textbf{94.9} & 69.3 & 52.9 \\ 
+GP (4k) & 68.0 & 38.8 & 25.9 & 30.9 & 85.8 & 65.8 & 52.5 \\ 
+PI (4k) & 68.3 & 36.0 & 25.9 & 30.8 & 84.9 & 65.3 & 51.9  \\ 
+SE (4k) & \textbf{73.5} & \textbf{53.5} & \textbf{32.3} & \textbf{39.1} & 91.9 & \textbf{74.6} & \textbf{60.8} \\
+NTK (4k) & 66.3 & 46.5 & 25.5 & 35.8 & 90.8 & 71.7 & 56.1 \\
\midrule
\textit{E5-Mistral} & \textit{71.0} & \textit{48.3} & \textit{44.6} & \textit{43.6} & \textit{96.8} & \textit{82.0} & \textit{64.4} \\ \midrule
+PCW (32k) & 63.5 & 49.5 & \textbf{59.3} & 51.3 & \textbf{97.3} & 91.2 & 68.7 \\
+GP (32k) & 81.0 & 48.8 & 37.0 & 42.9 & 90.6 & 88.1 & 64.7 \\ 
+PI (32k) & 89.8 & 48.5 & 37.8 & 40.4 & 76.8 & 63.0 & 59.4 \\ 
+SE (32k) & 90.8 & 52 & 49.3 & 48.7 & 97.2 & \textbf{96.4} & 72.4 \\
+NTK (32k) & \textbf{93.8} & \textbf{66.8} & 49.8 & \textbf{49.2} & 97.1 & 95.2 & \textbf{75.3} \\
\bottomrule
\end{tabular}
\caption{Results (\%) of context window extension methods on \efrbase{} and E5-Mistral. For datasets, \textit{P}, \textit{N}, \textit{NQA}, \textit{QMS}, \textit{SFD}, \textit{WQA} is short for \textit{Passkey}, \textit{Needle},  \textit{NarrativeQA}, \textit{QMSum}, \textit{SummScreenFD}, \textit{2WikiMultihopQA}. For extension methods, \textit{PCW}, \textit{GP}, \textit{PI}, \textit{SE}, \textit{NTK} are short for \textit{Parallel Context Windows}, \textit{Grouped Positions}, \textit{Linear Position Interpolation}, \textit{SelfExtend}, and \textit{NTK-Aware Interpolation}, respectively.}
\label{tab:rope_results}
\end{table}

\noindent \textbf{RoPE-based Models.}\quad %
Table~\ref{tab:rope_results} depicts the outcomes of \efrbase{} and E5-Mistral on each dataset of \benchmarkname{} after context window extension via PCW, GP, PI, SE and NTK. It is observed that RoPE-specific methods including NTK and SE yield significant improvements for both models across all datasets, surpassing PCW, PI and GP by a large margin.

\subsection{Analysis}
\label{sec:analysis}

\noindent \textbf{Tuning on PI vs. RP.}\quad
Figure~\ref{fig:pi_vs_rp} compares further tuning on top of RP vs. PI. 
In the former approach, the initial 512 position embeddings are frozen while the remaining embeddings are tuned, whereas for the latter, the frozen / learnable embedding vectors are arranged in an interleaved manner. 
We observe that tuning on PI consistently produces superior results on both \gtebase{} and \efbase{}. A possible explanation is that fixed vectors in PI serve intrinsically as anchors, preventing the learnable vectors from converging to suboptimal values.


\noindent \textbf{RoPE vs. APE.}\quad We further discuss the potential of APE / RoPE-based models for context window extension. \efbase{} and \efrbase{} are selected as the comparison subjects thanks to their shared training process, training data, and comparable performance on BEIR and \benchmarkname{} benchmarks. At each target context length ($\{1k, 2k, 4k\}$), we report the best scores achieved by each model on \benchmarkname{}, as illustrated in Figure~\ref{fig:rope_vs_ape}. Without requiring further training, \efrbase{} consistently demonstrates superior performance compared to \efbase{} across all target lengths. Furthermore, as the target window length increases, this superiority becomes more pronounced, even surpassing the fine-tuned version of \efbase{} by a large margin. This suggests that RoPE-based models can better extrapolate to to longer context. Consequently, we advocate for the use of RoPE in future embedding models.

\section{Conclusion}
This paper explores context window extension of existing embedding models. 
Through extensive experiments on our \benchmarkname{} benchmark, we show that training-free context window extension strategies can effectively increase the input length of these models by several folds. Further, our analysis reveals the superiority of RoPE-based embedding models over APE-based ones in context window extension. Hence, we advocate for the use of RoPE for future embedding models.

\section*{Limitations}
As a pioneering work in applying context window extension on embedding models, this paper is still limited in several aspects, particularly in that most of the context extension strategies explored in this paper are training-free. As evidenced by previous findings~\citep{xiong2023effective,fu2024data,zhang2024extending,yen2024longcontext}, and the additional performance gain achieved via tuning on \efbase{} and \gtebase{}, we believe
further fine-tuning on top of plug-and-play methods can bring even better extension results. In the future, we will make comprehensive exploration of training-based context window extension for embedding models, especially for RoPE-based ones.

\section*{Ethics Statement}
This work fully complies with the ACL Ethics Policy.
We declare that there are no ethical issues in this paper, to the best of our knowledge.

\section*{Acknowledgement}
We thank the anonymous reviewers for their helpful comments on this paper. We thank Xueguang Ma, Niklas Muennighoff, and Kenneth Enevoldsen for their thoughtful discussion and assistance in integrating LongEmbed into MTEB. This work was partially supported by National Natural Science Foundation of China (No. 62476010). 

\bibliography{bib/custom}
\bibliographystyle{bib/acl_natbib}

\appendix

\clearpage

\section{Training Details for \efrbase{}}
\label{sec:app_train_e5rope}

\begin{table}[h]
\centering
\footnotesize
\renewcommand{\arraystretch}{1.1}
\setlength\tabcolsep{1pt}
\begin{tabular}{l|cccc}
\toprule
\multirow{2.5}{*}{\textbf{Params}} &
\multicolumn{2}{c}{\textbf{Pre-training}} &
\multicolumn{2}{c}{\textbf{Fine-tuning}} \\
\cmidrule(r){2-3} \cmidrule(r){4-5}
& \efbase{} & \efrbase{} & \efbase{} & \efrbase{}  \\
\midrule
learning rate & 2$\times 10^{-4}$ & 2$\times 10^{-4}$ & 2$\times 10^{-5}$ & 2$\times 10^{-5}$ \\
GPUs (V100) & 32 & 32 & 8 & 8 \\
warmup steps & 1000 & 1000 & 400 & 400 \\
max length & 128 & 512 & 192 & 192 \\
batch size & 32k & 16k & 256 & 256 \\
max steps & 20k & 20k & n.a. & n.a. \\
epochs & n.a. & n.a. & 3 & 3 \\
$\tau$ & 0.01 & 0.01 & 0.01 & 0.01 \\
$\alpha$ & n.a. & n.a. & 0.2 & 0.2 \\
weight decay & 0.01 & 0.01 & 0.01 & 0.01 \\
hard negatives & 0 & 0 & 7 & 7 \\
pos embedding & APE & RoPE & APE & RoPE \\
\bottomrule
\end{tabular}
\caption{Hyperparameters for contrastive pre-training and fine-tuning of \efbase{} and \efrbase{}. }
\label{tab:param}
\end{table}

In this section, we describe the training details of \efrbase{}. Our training procedure and data exactly follows that of E5~\citep{wang2022text}, where we first perform contrastive pre-training on their collected CCPairs, then perform fine-tuning on the concatenation of 3 datasets: MS-MARCO passage ranking~\citep{nguyen2016ms}, NQ~\citep{karpukhin2020dense,kwiatkowski2019natural}, and NLI~\citep{gao2021simcse}. Each example is paired with 7 hard negatives. We leverage the mined hard negatives and re-ranker scores from SimLM~\citep{wang2023simlm} for the first two datasets. As the NLI dataset only provides 1 hard negative per example, we randomly sample 6 sentences from the entire corpus. xFormers~\citep{xFormers2022} is used for memory efficient training. As presented in Table~\ref{tab:param}, training hyperparameters for \efbase{} and \efrbase{} are identical, except in two aspects:
\begin{itemize}[leftmargin=*]
    \item \textbf{Initialization.}\quad Before contrastive pre-training, \efbase{} is initialized on BERT\textsubscript{Base}~\citep{devlin-etal-2019-bert}, which employs absolute position embeddings (APE). For the initialization of \efrbase{}, we simply replace the APE part of BERT\textsubscript{Base} with RoPE. It's worth noting that the BERT\textsubscript{Base} model after this replacement cannot function properly. We count on the subsequent pre-training phase to adapt the model to RoPE.
    \item \textbf{Pre-training length and batch size.}\quad \efbase{} does not update its position embedding matrix during the training phase, i.e., it utilizes the same position embedding matrix as BERT\textsubscript{Base}. This allows it to generalize to input sequences of up to 512 tokens, while being trained with a max training length of 192. As for E5-RoPE, replacing APE with RoPE during initialization prevents us from directly inheriting the original model's capability in handling 512 tokens. Consequently, in the pre-training phase of E5-RoPE, we set the maximum training length to 512, and reduce the batch size to 16k according to memory constraints.
\end{itemize}

\begin{table}[t]
\centering
\footnotesize
\renewcommand{\arraystretch}{1.1}
\setlength\tabcolsep{4pt}
\begin{tabular}{@{}lcccc@{}}
\toprule
\textbf{Tasks} & \textbf{\# W/Q.} & \textbf{\# W/D.} & \textbf{\efbase{}} & \textbf{\efrbase{}} \\ 
\midrule
MS MARCO & 6.0 & 56.0 & 41.8 & 42.4 \\
Trec-Covid & 10.6 & 160.8 & 69.6 & 73.3 \\
NFCorpus & 3.3 & 232.3 & 35.4 & 34.9 \\
NQ & 9.2 & 78.9 & 58.2 & 60.1 \\
HotpotQA & 17.6 & 46.3 & 69.1 & 61.0 \\
FiQA & 10.8 & 132.3 & 39.8 & 36.4 \\
ArguAna & 193.0 & 166.8 & 44.6 & 54.2 \\
Touche-2020 & 6.6 & 292.4 & 26.4 & 26.6 \\
CQADupStack & 8.6 & 129.1 & 37.4 & 36.5 \\
Quora & 9.5 & 11.4 & 86.6 & 87.7 \\
DBPedia & 5.4 & 49.7 & 42.2 & 40.0 \\
Scidocs & 9.4 & 176.2 & 18.7 & 18.1 \\
Fever & 8.1 & 84.8 & 85.0 & 68.0 \\
Climate-Fever& 20.1 & 84.8 & 26.6 & 19.0 \\
Scifact & 12.4 & 213.6 & 72.0 & 71.0 \\ \midrule
Average & < 200 & < 300 & 50.23  & 48.61 \\ \bottomrule
\end{tabular}
\caption{Statistics and performance comparison of \efbase{} and \efrbase{} on 15 publicly available BEIR tasks. \# W/Q. and \# W/D. stands for word number per query and per document, respectively. }
\label{tab:beir}
\end{table}

Table~\ref{tab:beir} demonstrates results of \efbase{} and \efrbase{} on 15 publicly available BEIR tasks. We observe comparable overall scores between both models. This comparable performance, along with their shared training process and training data, facilitates fair comparison of APE and RoPE-based models's capabilities in length extrapolation. Note that the slight performance loss of \efrbase{} could possibly be attributed to the replacement of position embedding in the initialization phase, or the reduced batch size in the pre-training phase, as mentioned before.

\begin{table*}[t]
\centering
\footnotesize
\renewcommand{\arraystretch}{1.1}
\setlength\tabcolsep{15pt}
\begin{tabular}{lccc}
\toprule
\textbf{Extension} & \textbf{PCW \& GP \& RP \& PI} & \textbf{NTK} & \textbf{SE} \\\midrule
\multicolumn{4}{c}{\textit{\gtebase{} \& \efbase{}}} \\ \midrule
512 -> 1,024 & $L_o=512,L_t=1,024,s=2$ & - & - \\
512 -> 2,048 & $L_o=512,L_t=2,048,s=4$ & - & - \\
512 -> 4,096 & $L_o=512,L_t=4,096,s=8$ & - & - \\ \midrule
\multicolumn{4}{c}{\textit{ \efrbase{}}} \\ \midrule
512 -> 1,024 & $L_o=512,L_t=1,024,s=2$ & $\lambda=3$~(10,000 -> 30,000) & $g=3, w=256$ \\
512 -> 2,048 & $L_o=512,L_t=2,048,s=4$ & $\lambda=5$~(10,000 -> 50,000) & $g=5, w=128$ \\
512 -> 4,096 & $L_o=512,L_t=4,096,s=8$ & $\lambda=10$~(10,000 -> 100,000) & $g=9, w=64$ \\ \midrule
\multicolumn{4}{c}{\textit{E5-Mistral}} \\ \midrule
4,096 -> 8,192 & $L_o=4,096,L_t=8,192,s=2$ & $\lambda=3$~(10,000 -> 30,000) & $g=3, w=2,048$ \\
4,096 -> 16,384 & $L_o=4,096,L_t=16,384,s=4$ & $\lambda=5$~(10,000 -> 50,000) & $g=5, w=1,024$\\
4,096 -> 32,768 & $L_o=4,096,L_t=32,768,s=8$ & $\lambda=10$~(10,000 -> 100,000) & $g=9, w=512$\\
\bottomrule
\end{tabular}
\caption{Hyperparameters for plug-and-play context extension strategies.}
\label{tab:extend_hyper}
\end{table*}

\section{Implementation Details for Context Extension Strategies}
\label{sec:app_implementation}
This section describes implementation details for the explored context extension stratgies. For plug-and-play methods including PCW, RP, GP, PI, NTK and SE, Table~\ref{tab:extend_hyper} summarizes their hyperparameters under each condition.

\noindent \textbf{Further Tuning.}\quad On top of PI and RP,  we perform further tuning on both \efbase{} and \gtebase{}, utilizing the fine-tuning dataset mentioned in Appendix~\ref{sec:app_train_e5rope}. Following the practice of PoSE~\citep{zhu2023pose}, we manipulate position ids to simulate long training samples. Concretely, given an input document $\mathcal{D}=\{x_0,x_1,...,x_{L_o-1}\}$ of original context length $L_o$, we introduce a skipping bias term $u$ at the beginning of $\mathcal{D}$, transferring the original position ids $\mathcal{D}$ into $\{0,1,...,L_o-1\}$  into $\{u,u+1,...,u+L_o-1\}$.~\footnote{The original practice of PoSE focuses on relative position, hence introduces bias terms at the middle of document $\mathcal{D}$. For APE-based models, we simply skips from the beginning.} For every piece of training data, $u$ is re-sampled from the discrete uniform distribution $\mathcal{U}(\{0,1,...,L_t-L_o\})$. In this way, we ensure comprehensive coverage of target context window.
The training procedure spans 3 epochs on 2 A100 GPUs, with a learning rate of $5e^{-4}$, a batch size of 512, and 100 steps for warmup. Other hyperparameters are same as Table~\ref{tab:param}.

\noindent \textbf{Inference.}\quad In inference time, attention scaling~\citep{kexuefm8823,chiang-cholak-2022-overcoming} is used by default for all tested models for better length extrapolation ability. Especially for \gtebase{} and \efbase{} tuned on PI, we use the original position ids when input length not exceeds 512. This is achived by mapping the position ids $\{0,1,...,l\}$ into $\{0,s,...,l\times s\}$, where $s$ is the scaling factor, $l< 512$.

\section{Further details on \benchmarkname{}}
\label{app:sec_longembed_details}

\begin{figure*}[t]
    \includegraphics[width=\linewidth]{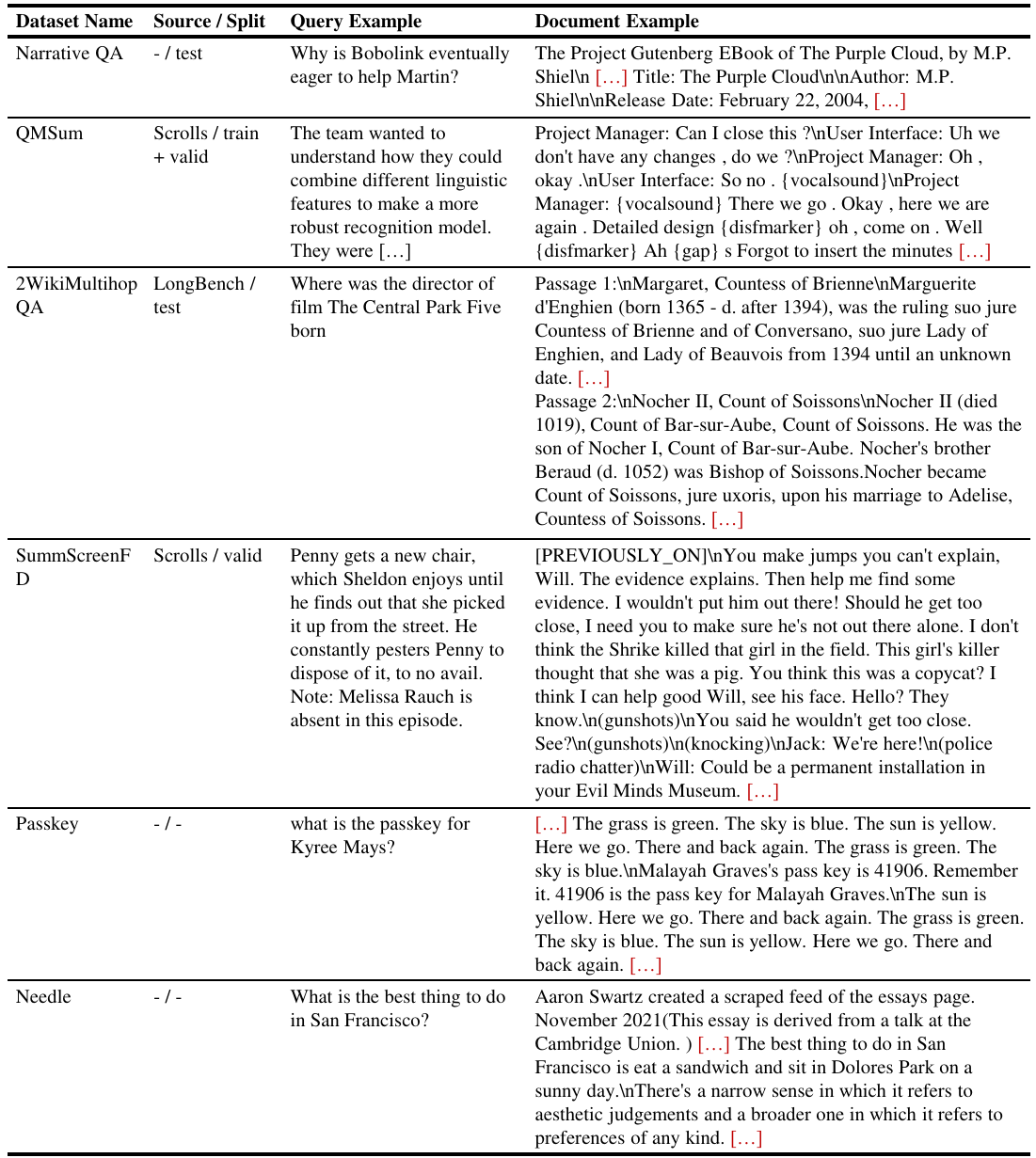}
    \caption{Source and examples for each dataset in \benchmarkname{}.}
    \label{fig:longembed_example}
\end{figure*}
Figure~\ref{fig:longembed_example} presents source and examples for each dataset included in \benchmarkname{}. 
For QA datasets including NarrativeQA and 2WikiMultihopQA, we adopt their test splits. Note that for 2WikiMultihopQA, we adopt the length-uniformly sampled version from ~\citet{bai2023longbench} to better assess the model’s capabilities across various context lengths.
For summarization datasets including QMSum and SummScreenFD, we adopt the version processed by SCROLLS~\citep{shaham-etal-2022-scrolls}. Since SCROLLS does not include ground truth summarization in its test sets, we switch to validation set for these two datasets. Particularly for QMSum, as its validation set only have 60 documents, which is too small for document retrieval, we included the train set as well.

\section{BM25 Results on \benchmarkname{}}

\begin{table}[t]
\centering
\footnotesize
\renewcommand{\arraystretch}{1.1}
\setlength\tabcolsep{2.5pt}
\begin{tabular}{lccccccc}
\toprule
\multirow{2.5}{*}{\textbf{Method}} &
\multicolumn{2}{c}{\textbf{Synthetic}} &
\multicolumn{4}{c}{\textbf{Real}} & \multirow{2.5}{*}{\textbf{Avg.}}  \\
\cmidrule(r){2-3} \cmidrule(r){4-7}
& P & N & NQA & QMS & SFD & WQA & \\
\midrule
BM25 & \textbf{100} & \textbf{95.3} & \textbf{71.5} & \textbf{81.3} & \textbf{97.6} & \textbf{96.5} & \textbf{90.4} \\ \midrule
E5-Mistral & 71.0 & 48.3 & 44.6 & 43.6 & 96.8 & 82.0 & 64.4 \\
+NTK (32k) & 93.8 & 66.8 & 49.8 & 49.2 & 97.1 & 95.2 & 75.3 \\
\bottomrule
\end{tabular}
\caption{BM25 Results on \benchmarkname{}. \textit{P}, \textit{N}, \textit{NQA}, \textit{QMS}, \textit{SFD}, \textit{WQA} is short for \textit{Passkey}, \textit{Needle},  \textit{NarrativeQA}, \textit{QMSum}, \textit{SummScreenFD}, \textit{2WikiMultihopQA}.}
\label{tab:bm25_results}
\end{table}
Table~\ref{tab:bm25_results} shows the scores of BM25 on \benchmarkname{}, along with those of the best-performing long context embedding model, E5-Mistral. The significant gap between BM25 and E5-Mistral highlights substantial room for improvement in current long context embedding models.

\end{document}